\documentclass{article}

\usepackage{microtype}
\usepackage{graphicx}
\usepackage{subcaption}
\usepackage{booktabs} % for professional tables
\usepackage{multirow}
\usepackage{rotating}
\usepackage{tikz}
\usepackage{threeparttable}
\usepackage{hyperref}

\usepackage[accepted]{icml2023}

\usepackage{amsmath}
\usepackage{amssymb}
\usepackage{mathtools}
\usepackage{amsthm}

\usepackage[capitalize,noabbrev]{cleveref}

\theoremstyle{plain}

\theoremstyle{definition}

\theoremstyle{remark}

\usepackage[textsize=tiny]{todonotes}

\icmltitlerunning{SimTS: Rethinking Contrastive Representation Learning for Time Series Forecasting}

\begin{document}

\twocolumn[
\icmltitle{SimTS: Rethinking Contrastive Representation Learning for Time Series Forecasting}



\icmlsetsymbol{equal}{*}

\begin{icmlauthorlist}
\icmlauthor{Xiaochen Zheng}{equal,uzh,ai}
\icmlauthor{Xingyu Chen}{equal,eth}
\icmlauthor{Manuel Schürch}{uzh,ai}
\icmlauthor{Amina Mollaysa}{uzh,ai}
\icmlauthor{Ahmed Allam}{uzh}
\icmlauthor{Michael Krauthammer}{uzh,ai}
\end{icmlauthorlist}

\icmlaffiliation{uzh}{University of Zurich, Zurich, Switzerland}
\icmlaffiliation{eth}{ETH Zurich, Zurich, Switzerland}
\icmlaffiliation{ai}{ETH AI Center, Zurich, Switzerland}

\icmlcorrespondingauthor{Xiaochen Zheng}{xiaochen.zheng@uzh.ch}

\icmlkeywords{Time series forecasting, representation Learning, predictive coding}

\vskip 0.3in
]



\printAffiliationsAndNotice{\icmlEqualContribution} % otherwise use the standard text.

\begin{abstract}
% While existing contrastive learning methods have shown an impressive ability to learn meaningful representations for image or time series classification, they face challenges when applied to time series forecasting, as (1) optimization of instance discrimination is not directly applicable to predicting the future state from the history context, and (2) construction of positive and negative pairs across diverse types of time series data is difficult.  SimTS overcomes these challenges by learning to predict future time series windows from the past in the latent space without relying on negative pairs or specific assumptions about the characteristics of the particular time series. Our extensive experiments on several benchmark time series forecasting datasets show that SimTS achieves competitive performance compared to state-of-the-art contrastive learning methods. Furthermore, we show the shortcomings of the current contrastive learning framework used for time series forecasting through a detailed ablation study. Overall, our work suggests that SimTS is a promising alternative to other contrastive learning approaches for time series forecasting.

Contrastive learning methods have shown an impressive ability to learn meaningful representations for image or time series classification. However, these methods are less effective for time series forecasting, as optimization of instance discrimination is not directly applicable to predicting the future state from the history context. Moreover, the construction of positive and negative pairs in current technologies strongly relies on specific time series characteristics, restricting their generalization across diverse types of time series data. To address these limitations, we propose SimTS, a simple representation learning approach for improving time series forecasting by learning to predict the future from the past in the latent space. SimTS does not rely on negative pairs or specific assumptions about the characteristics of the particular time series. Our extensive experiments on several benchmark time series forecasting datasets show that SimTS achieves competitive performance compared to existing contrastive learning methods. Furthermore, we show the shortcomings of the current contrastive learning framework used for time series forecasting through a detailed ablation study. Overall, our work suggests that SimTS is a promising alternative to other contrastive learning approaches for time series forecasting.

\end{abstract}
 
\section{Introduction}
The field of time series forecasting has experienced significant progress in recent years with a wide range of practical applications across different sectors such as finance~\cite{sezer2020financial}, traffic ~\cite{traffic}, and clinical practice~\cite{mimic}.  The availability of large volumes of data is one of the key factors behind these advancements. In particular, self-supervised learning approaches such as contrastive learning ~\cite{yue2022ts2vec,woo2022cost,ncl} have shown promise in exploiting these datasets and have continually outperformed supervised approaches~\cite{CNN,salinas2020deepar,informer} in time series forecasting tasks. Self-supervised contrastive approaches learn representations by mapping similar instances (i.e., positive pairs) to similar representations while pushing dissimilar instances (i.e., negative pairs) apart. Most contrastive learning approaches rely on instance discrimination~\cite{instance_dis}. The resulting representations contain information that can discriminate well between different instances of time series, making them informative for downstream tasks such as time series \textit{classification}. However, in time series \textit{forecasting}, the goal is to predict the future based on past time windows rather than discriminating between instances. Consequently, features learned by instance discrimination may not be sufficient for accurate forecasting. 

\begin{figure}[!t]
    \includegraphics[width=0.3\textwidth]{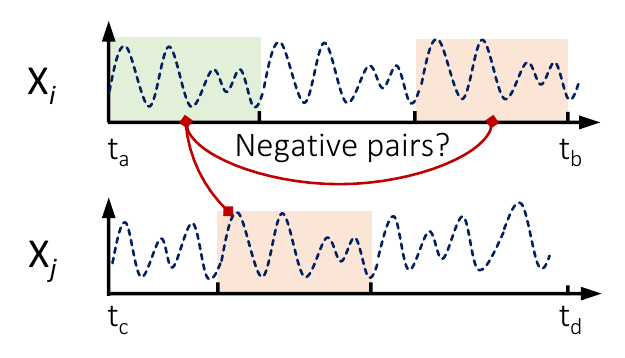}
    \centering
    \caption{Problems with selecting negative pairs based on methods proposed in~\cite{ncl, yue2022ts2vec, woo2022cost} when cross-instance and cross-time repeated patterns exist.}
    \label{fig:falseneg}
\end{figure}

Additionally, identifying positive and negative pairs for time series forecasting is challenging. Contrastive learning relies on data augmentations to generate positive pairs. While it is possible to find semantic preserving augmentations for time series classification~\cite{shapelet,ncl,expertclr}, it is more difficult to identify augmentation methods that can be generalized to time series forecasting. Besides, most existing methods for constructing negative pairs depend heavily on the individual characteristics of the time series, making them not applicable to other types of time series. \citet{yue2022ts2vec,clocs,ncl,ICA,tonekaboni2021tnc} propose methods based on the assumptions that (1) the similarity between segments of the same time series decreases as the time lag increases, and (2) segments of distinctive time series are dissimilar. However, particular time series do not adhere to these assumptions, resulting in unsatisfactory representations~\cite{tstcc,expertclr} in other time series. For instance, in a time series with a strong periodicity, similar patterns exist between or within instances. As illustrated in Figure~\ref{fig:falseneg}, selecting times-windows randomly may result in selecting inappropriate negative pairs~\cite{goodviews}, leading to false repulsion, where the model incorrectly discriminates representations of similar samples. Other recent approaches are based on disentanglement ~\cite{woo2022cost,wang2022last} or fusion ~\cite{yang22ebilinear,tfc}, assuming that a time series can be represented by trends and seasonality components. As a result, these approaches may not generalize well across various forecasting datasets, since real-world data often lack consistent seasonality. In general, this reliance on specific characteristics limits their generalizability when applied to different types of time series data, which will be demonstrated through detailed experiments in Section~\ref{sec:exp}.

To address these limitations in contrastive representation learning for time series forecasting, the paper aims to answer the following key question: ``what is important for time series forecasting with contrastive learning, and how can we adapt contrastive ideas more effectively to time series forecasting tasks?” Beyond contrastive learning, we propose a \textit{\textbf{Sim}ple Representation Learning Framework for \textbf{T}ime \textbf{S}eries Forecasting} (\textbf{SimTS}), which is inspired by predictive coding~\cite{cpc}: we learn a representation such that the latent representation of the future time windows can be predicted from the latent representation of the history time windows. In particular, we build upon a siamese network structure~\cite{siamese,simsiam} and propose key refinements that enable better prediction performance with a simpler model structure compared to state-of-art methods. First, we divide a given time series into \textit{history} and \textit{future} segments and then use an encoding network to map them to their latent space. Second, we use a predictive layer to predict the latent representation of the \textit{future} segment from the \textit{history} segment. We regard the predicted representation (from the \textit{history} segment) and the encoded representation of the \textit{future} segment as positive pairs. The representations learned in this way encode features that are useful for forecasting tasks.

Moreover, the paper questions existing assumptions and techniques used for constructing positive and negative pairs. We provide a detailed discussion and several experiments showing their shortcomings when applied to various time series. Specifically, inspired by~\cite{simsiam,byol,yuandong_pos}, we question the proposed usage of negative pairs for time series forecasting and the idea of augmenting the data to generate positive pairs, which is empirically investigated in several experiments with different contrastive methods. As a consequence, our model does not use negative pairs to avoid false repulsion. We hypothesize that the most important mechanism behind representation learning for time series forecasting is maximizing the shared information between representations of \textit{history} and \textit{future} time windows. In our proposed model, we explicitly impose a constraint that the learned representation of history should encode as much information as possible by predicting the latent representation of the future from the latent representation of history. This mechanism simplifies several existing approaches and leads to state-of-the-art forecasting results, as thoroughly demonstrated in this paper.

Our contributions can be summarized as follows:

%\begin{itemize}
%    \item We conduct extensive experiments to assess and evaluate the effectiveness of various assumptions that are widely used in current state-of-the-art contrastive learning frameworks. We propose SimTS, demonstrating it outperforms the existing frameworks while requiring no negative pairs on benchmark tasks.
%    \item We propose a novel way of designing positive pairs instead of using data augmentation. Our model leverages the latent representation of history to anticipate the latent representation of the future, which compels the model to acquire an effective representation for forecasting purposes.
%    \item We develop a simple yet efficient encoder that can capture temporal dependencies at various scales. We construct a combination of convolution layers with diverse kernel size (i.e. receptive field) by choosing the appropriate look-back window. 
%\end{itemize}

\begin{itemize}
    \item We propose a novel method (SimTS) for time series forecasting, which employs a siamese structure and a simple convolutional encoder to learn representations in latent space without requiring negative pairs.
    \item We demonstrate the effectiveness and generalizability of SimTS across various types of time series through experiments on multiple types of benchmark datasets. Our method outperforms state-of-the-art methods for multivariate time series forecasting.
    \item We conduct extensive ablation experiments to assess and evaluate the effectiveness of various assumptions that are widely used in current state-of-the-art contrastive learning frameworks. This provides insights into the key factors that contribute to the performance of time series forecasting and sheds light on potential areas for improvement in future research.
\end{itemize}
\section{Related Works}
\label{relatedWork}
%Time series forecasting (end to end)
Researchers have recently developed numerous deep learning models to address the challenges of time series forecasting. Traditional models for time series prediction, such as ARIMA~\cite{arima}, SVM~\cite{svm}, and VAR~\cite{VAR}, have been outperformed on many datasets by deep learning models, including RNN~\cite{RNN}, CNN~\cite{CNN} and transformers~\cite{transformer}. TCN~\cite{CNN} introduces dilated convolutions~\cite{wavenet} for time series forecasting, which incorporates dilation factors into conventional CNNs to increase the receptive field significantly. To improve the effectiveness of long-term time series forecasting, the conventional transformer is modified and applied to time series: LogTrans~\cite{logTrans} suggests the \textit{LogSparse} attention; Informer~\cite{informer} develops the \textit{ProbSparse} self-attention mechanism to reduce the computational cost of long-term forecasting. 

%unsupervised time series methods-->decomposition
Recent developments in self-supervised learning have successfully discovered meaningful representations for images~\cite{moco,simclr} with InfoNCE loss~\cite{cpc}. To get reliable time-series representations, several approaches have been investigated. Some studies focus on formulating time segments as contrastive pairs:  ICA~\cite{ICA} investigates non-stationarity in temporal data to find a representation that allows optimal time segment discrimination; TNC~\cite{tonekaboni2021tnc} establishes a temporal neighborhood to contrast between neighboring segments and learn the underlying temporal dependency of non-stationary time series. However, these methods do not perform as well in forecasting tasks since they focus on extracting neighborhood features and fail to capture global patterns in time series. Furthermore, some methods utilize more complicated contrastive learning approaches to learn effective representations for time series. For example, \cite{franceschi2019unsupervised} learns scalable representations for various time series lengths using contrasting positive, negative, and reference pairs with an innovative triplet loss. TS2Vec~\cite{yue2022ts2vec} employs hierarchical contrastive learning over time series augmentations, generating  representations for each time step. However, these approaches formulate contrastive learning frameworks as classification tasks, which try to learn representations by discriminating time series from different classes and therefore ignore learning predictive features. Additionally, as time series can be (re-)constructed by combining trend, season, and noise components~\cite{decomposition}, there is growing research that uses time series decomposition in unsupervised learning. CoST~\cite{woo2022cost} encodes disentangled trend and seasonal representations using contrastive learning. BTSF~\cite{yang22ebilinear} aggregates time and spectral domain to extract global information and refine representations. While decomposition-related methods may exhibit robust performance in certain datasets, they heavily rely on underlying assumptions about the data's characteristics and tend to fail when dealing with datasets that lack specific seasonality or trend.
\section{Methods}

\subsection{Motivation}

In this work, we rethink ``what is important for time series forecasting with contrastive learning?" Firstly, we observe that the existing methods might (1) ignore the possibility that repeated patterns exist within a time series, even though they may be located far apart from one another, and (2) disregard the possibility that distinct time series may contain similar patterns. We aim to identify a more suitable design that considers the inherent nature of time series forecasting and adheres to necessary assumptions for effective representation learning. We argue that a good representation should effectively capture the temporal dependencies between past segments and future predictions in forecasting tasks, emphasizing that the temporal differences hold greater significance than the similarity between positive and negative pairs. Thus, we design predictive positive pairs that can learn more flexible and adaptive representations.

Secondly, current approaches require sufficient negative pairs to avoid collapsing~\cite{simclr,simsiam,collapse}. Collapsing happens in Siamese networks~\cite{siamese} where the model produces a constant representation regardless of the input. Although the introduction of negative pairs constrains the solution space and prevents collapsing, it might also induce the issue of false repulsion. Simultaneously, identifying suitable augmentation methods and negative pairs for forecasting tasks can be challenging, especially when repeated patterns exist across different samples. Such challenges motivate us to explore alternative approaches that circumvent negative pairs and implement stop-gradient solutions.

Furthermore, we contend that real-world data often lack distinct seasonality, making it difficult for models to learn irregular temporal information using abstract features. Our experiments demonstrate that learned representations, which discard some additional model components, yield better forecasting performance than the state-of-the-art contrastive model CoST~\cite{woo2022cost}, as shown in Table~\ref{tab:disentangle}. These results suggest that current methods may not generalize well to diverse time series datasets. Finally, it leads us to the central motivation of our SimTS model: we train an encoder to learn time series representations by predicting its future from historical segments in the latent space. SimTS achieves the best performance in time series forecasting benchmark datasets with a relatively simpler design compared to other contrastive learning frameworks.

\subsection{SimTS: Simple Representation Learning for Time Series Forecasting}
Given a time series $X =[x_{1},x_{2},\dots,x_{T}] \in\mathbb{R}^{C\times T}$, where $C$ is the number of features (i.e., variables) and $T$ denotes the sequence length. Our objective is to learn a latent representation of the \textit{history} segment $X^h=[x_{1},x_{2},\dots,x_{K}]$, where $0<K<T$, such that our model can predict the \textit{future} segment $X^f=[x_{K+1},x_{K+2},\dots,x_{T}]$ from it. 

Inspired by well-developed contrastive learning frameworks~\cite{cpc,simclr,byol,simsiam}, SimTS learns time series representations by maximizing the similarity between predicted and encoded latent features for each timestamp. The approach involves designing an encoder network, denoted as $F_\theta$, which maps historical and future segments to their corresponding latent representations, $Z^h$ and $Z^f$, respectively. The encoder's objective is to learn an informative latent representation $Z^h = F_\theta(X^h) = [z^h_1, z^h_2, ..., z^h_K] \in \mathbb{R}^{C'\times K}$ that can be used to predict the latent representation of the future through a prediction network. The SimTS model consists of four main parts:

\begin{itemize}
    \item A siamese neural network architecture~\cite{siamese,simsiam} consisting of two identical networks that share parameters. The time series is divided into the \textit{history} segment $X^h$, and \textit{future} segment $X^f$, and given as inputs to the siamese network. The siamese network learns to map them to their latent representations $Z^h, Z^f$. %is a type of neural network architecture used to determine input similarity. It
    \item A multi-scale encoder consisting of a projection layer that projects raw features into a high dimensional space and multiple CNN blocks with different kernel sizes.%adaptive look-back windows.
    \item A predictor network $G_\phi$ that takes the last column of the encoded \textit{history} view as input and predicts the \textit{future} in latent space. 
    \item A cosine similarity loss that only takes positive samples into account.
\end{itemize}

\begin{figure}[!ht]
    \includegraphics[trim={0.3cm 0 0.3cm 0},clip,width=0.4\textwidth]{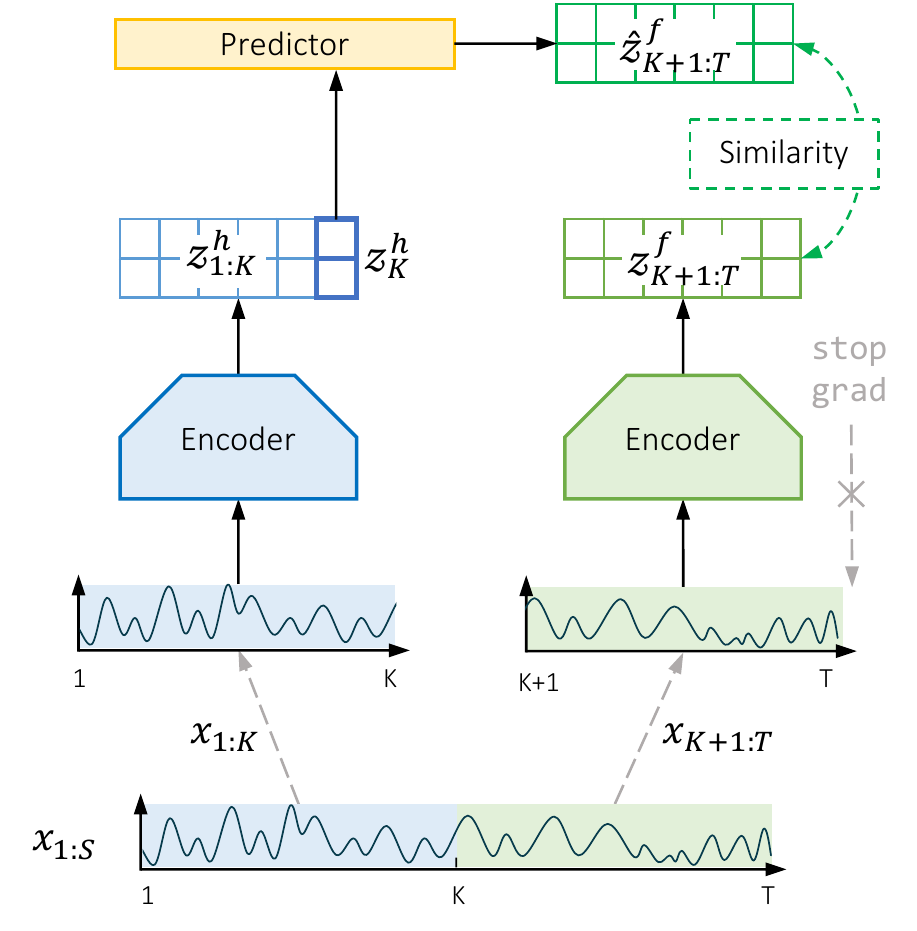}
    \centering
    \caption{Illustration of our proposed SimTS.}
    \label{fig:model}
\end{figure}

Figure~\ref{fig:model} depicts the overall architecture of SimTS. Our model architecture consists of two paths: the \textit{history} encoding path and the \textit{future} encoding path. The \textit{history} encoding path takes the \textit{history} view $X^h$ and outputs $Z^h = F_\theta(X^h)$. The \textit{future} encoding path takes the \textit{future} view $X^f$ and outputs the encoded latent representation of the future $Z^f = F_\theta(X^f)= [z^f_{K+1}, z^f_{K+2}, ..., z^f_{T}] \in \mathbb{R}^{C'\times (T-K)}$. As proposed in~\cite{byol,linearPred}, we apply a predictive MLP network $G_\phi$ on the last column of $Z^h$, denoted as $z^h_K$, to predict the \textit{future} latent representations: $\hat{Z}^f = G_\phi(z^h_K) = [\hat{z}^f_{K+1}, \hat{z}^f_{K+2}, ..., \hat{z}^f_{T}] \in \mathbb{R}^{C'\times (T-K)}$. Intuitively, the last column allows the encoder to condense the history information into a summary by properly choosing the kernel size. The training objective is to attract the \textit{predicted} \textit{future} and \textit{encoded }\textit{future} timestamps in representation space without introducing the negative pairs. As the predicted future latent representation is learned from the latent representation of the history, by forcing the predicted latent representation of the future to be close to the encoded latent representation of the future, we are forcing the model to learn a representation of the history that is informative for the future. Therefore, we regard the encoded $Z^f$ and the predicted \textit{future} representations $\hat{Z}^f$ as the positive pair and calculate the negative cosine similarity between them:

\begin{equation}
\label{eqn:sim}
Sim(\hat{Z}^f, Z^f) = -\frac{1}{T-K}\sum_{i=K+1}^{T} \frac{\hat{z}^f_i}{\parallel\hat{z}^f_i\parallel_2}\cdot \frac{z^f_i}{\parallel z^f_i\parallel_2}, 
\end{equation}
where $\parallel\cdot\parallel_2$ is $l_2$-norm and $Sim(\cdot)$ is the average cosine similarity of all time steps. Algorithm~\ref{alg:SimTS} summarises the proposed SimTS. 
\subsection{Multi-Scale Encoder}

\begin{figure}[!ht]
    \includegraphics[width=0.48\textwidth]{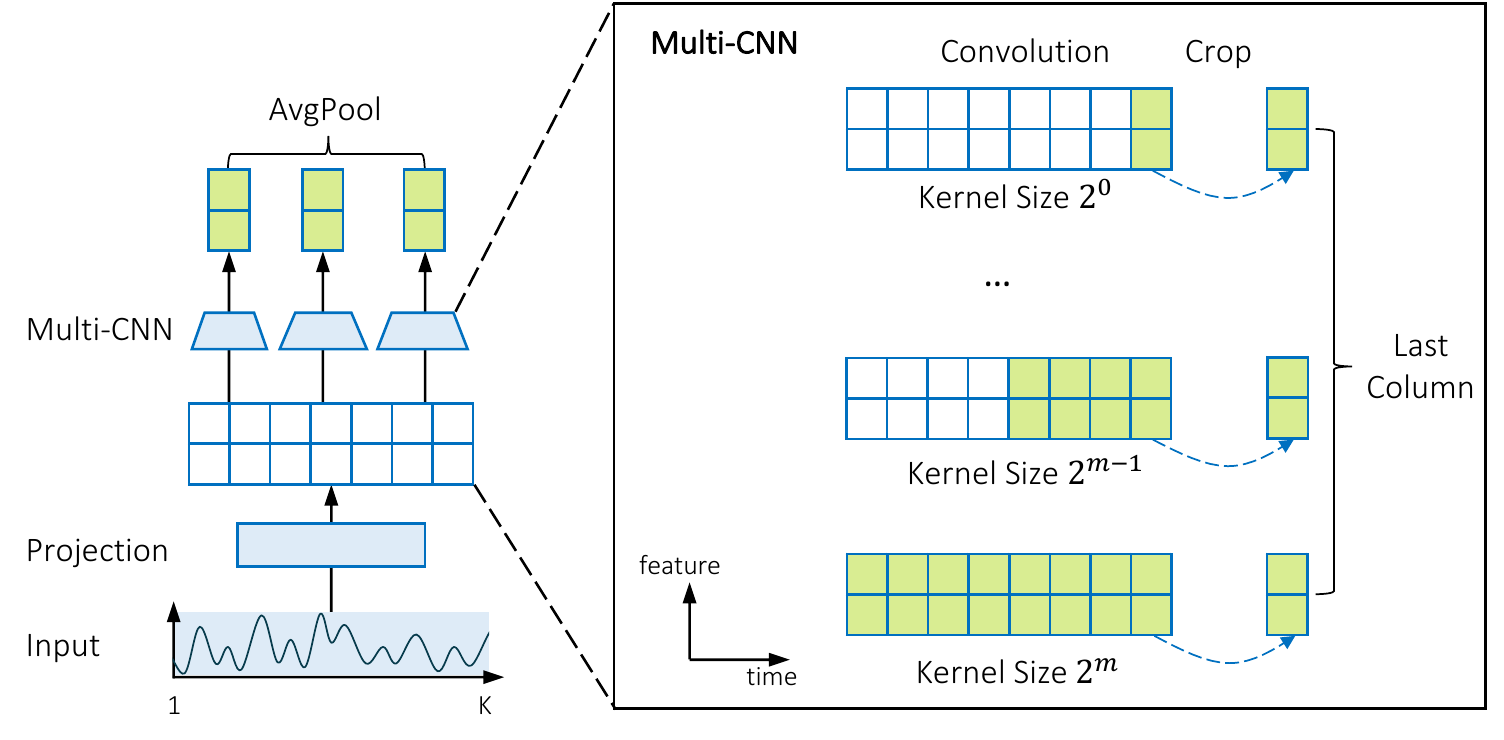}
    \centering
    \caption{Multi-scale encoder. Composed of a projection layer and a set of parallel 1d convolutions with kernel size $2^i$, for $i \in \{0,1,...,m\}$. An averaged pooling layer is added on the top of convolutions.}
    \label{fig:encoder}
\end{figure}
To learn a meaningful representation, the structure of the encoder network $F_\theta$ plays a vital role. Given the nature of time series, we would like our base encoder $F_\theta$ to extract temporal (inter-time) dependency from local and global patterns. For short-term forecasting, shorter local patterns (i.e., motifs) are ideal, whereas, for long-term forecasting, longer sets of global patterns are preferred. Therefore, we propose to use a convolutional network with multiple filters that have various kernel sizes, which can extract both global and local patterns.

Figure~\ref{fig:encoder} illustrates the details of the encoder $F_\theta$. First, each time series input is passed through a convolutional projection layer. The projection layer enables us to project time series into a latent space~\cite{yue2022ts2vec,woo2022cost,wang2022last}. We aim to capture abstract information and consistent intra-time relationships between features that may not be immediately apparent from the raw data. So that the model can potentially learn more informative and abstract representations of the raw inputs. Second, for a time series $X$ with length $K$, we have $m=\left [ log_2K \right ]+1$ parallel convolution layers on the top of the projection layer, and the $i$th convolution has kernel size $2^i$, where $i \in \{0,1,...,m\}$. These different kernel sizes can extract corresponding local/global patterns. Each convolution $i$ takes the latent features from the projection layer and generates a representation $\hat{Z}_{(i)}$. The final multi-scale representation $Z$ are obtained by averaging across $\hat{Z}_{(0)},\hat{Z}_{(1)},...,\hat{Z}_{(m)}$.

\begin{algorithm}[tb]
   \caption{SimTS's PyTorch-like Pseudocode}
   \label{alg:SimTS}
    \begin{algorithmic}
   %\STATE {\bfseries Input:} time series $X$, size $L\times C$
   \STATE initialize $\theta, \phi$
   \STATE given a mini-batch $\mathcal{D}=\{X_i\}_{i\in[1:N]}$ with N samples 
   \FOR{$X$ {\bfseries in $\mathcal{D}$}}
   \STATE $X^h$, $X^f$ = $X\left [:, :K, : \right ]$, $X\left [:, K:, : \right ]$\\
   \STATE $Z^h, Z^f = F_\theta(X^h), F_\theta(X^f)$
   \STATE $\hat{Z}^f = G_\phi(Z^h [ : , K , : ]$)
   \STATE $\hat{Z}^f$ = normalize($\hat{Z}^f$)
   \STATE $Z^f$ = normalize($Z^f$).detach()
   \STATE $L$ = - $(\hat{Z}^f\cdot Z^f)$.mean()
   \STATE $L$.backward() 
   \STATE update($F_\theta,G_\phi$)
   \ENDFOR
   \end{algorithmic}
\end{algorithm}

\subsection{Stop-gradient Operation }
 
We apply a stop-gradient operation~\cite{simsiam} to the \textit{future} encoding path in our model. Considering that we learn to encode both \textit{history} and \textit{future} using the same encoder, the model may optimize the encoder by pushing encoded \textit{future} $Z^f$ towards the predicted \textit{future} $\hat{Z}^f$.  As the encoder should constrain the latent of the past to be predictive of the latent of the \textit{future}, only $\hat{Z}^f$ can only move towards $Z^f$ in the latent space, not vice versa~\cite{collapse}. Due to the stop-gradient operation on $Z^f$, our encoder cannot receive updates from \textit{future} representations $Z^f$ and is constrained to only optimize the \textit{history} representation and its prediction $\hat{Z}^f$. With stop-gradient (\texttt{sg}), the loss is:

\begin{equation}
\begin{array}{l@{}l}
\label{eqn:loss}
\mathcal{L}_{\theta, \phi}(X^h, X^f) 
&\ = Sim\left(
G_\phi\left(
F_\theta(X^h)
\right), 
F_{\texttt{sg}(\theta)}(X^f)
\right)\\
&\ = Sim(\hat{Z}^f, \texttt{sg}(Z^f))
\end{array}
\end{equation}
The loss in definition \eqref{eqn:loss} is for one sample 
$X = [X^h,X^f]$. The loss for a mini-batch $\mathcal{D}=\{X_i^h,X_i^f\}_{i\in[1:N]}$ can be written as
\begin{equation}
% \tilde{
\mathcal{L}_{\theta, \phi}(\mathcal{D}) 
=
\frac{1}{N}
\sum_{i=1}^N
\mathcal{L}_{\theta, \phi}(X_i^h, X_i^f),
\end{equation}
which corresponds to the average loss across all samples in the mini-batch.
\section{Experiments}
\label{sec:exp}
As our goal is to learn a meaningful representation for various types of time series data for forecasting tasks, we focus on experimental settings where we can test the representation power of our model on various forecasting benchmark datasets. To keep a fair comparison, we follow the exact same setup as in CoST and TS2Vec. We first use our trained model to obtain the latent representation of the time series, then train a ridge regression model on the learned latent representation for forecasting, i.e., predicting future $L$ time steps. 
\subsection{Datasets and Baselines} 

 We compare our method to the most recent state-of-the-art two-stage representation learning methods for time series as well as to end-to-end learning methods where the model includes both the representation learning part and forecasting part are trained in an end-to-end fashion. 
The representation learning approaches include TS2Vec~\cite{yue2022ts2vec}, CoST~\cite{woo2022cost} and TNC~\cite{tonekaboni2021tnc} and end-to-end models include Informer~\cite{informer} and LogTrans~\cite{logTrans}, and two-stage representation learning approaches include TS2Vec, CoST and TNC. The details and implementations of the baselines are provided in the appendix. Our model was tested for both univariate and multivariate forecasting.  In the case of a dataset with $C$ features, we either predict the future values for all $C$ features (i.e., multivariate forecasting) or only focus on forecasting the future values of one specific feature (univariate forecasting).

Our experiments are carried out on six real-world public benchmark datasets. \textbf{Electricity Transformer Temperature (ETT)}~\cite{informer} measures long-term deployment of electric power. It consists of two hourly-sampled datasets (ETTh) and two 15-minute-sampled datasets (ETTm), which are collected for 2 years and from two different Chinese provinces.  ETT datasets contain one oil temperature feature and six power load features. In univariate forecasting, we only take oil temperature to train and forecast. In multivariate forecasting, we employ all features in our training and prediction. 
\textbf{Exchange-Rate}\footnote{\href{https://github.com/laiguokun/multivariate-time-series-data}{https://github.com/laiguokun/multivariate-time-series-data}}~\cite{exchangerate} contains the daily exchange rates of eight foreign countries from 1990 to 2016, including Australia, Britain, Canada, Switzerland, China, Japan, New Zealand, and Singapore. We consider the values of Singapore for univariate forecasting and all countries' value for multivariate forecasting.
\textbf{Weather}\footnote{\href{https://www.bgc-jena.mpg.de/wetter/}{https://www.bgc-jena.mpg.de/wetter/}} consists of local climatological data for almost 1,600 U.S. areas for 4 years. The data is collected every 10 minutes. Each time step contains 11 weather variables and one target feature, `Wet Bulb Celsius.' In univariate forecasting, we only consider the feature `Wet Bulb Celsius'; in multivariate forecasting, all features are included. The detailed statistics of the datasets are in the appendix (Table~\ref{tab:dataset}).

\begin{table*}[!htt]
    \centering
    \resizebox{\textwidth}{!}{
        % Table generated by Excel2LaTeX from sheet 'Main Results'
\begin{tabular}{cccccccccccccc}
\toprule
\multicolumn{1}{c}{\multirow{1}[2]{*}{Methods}} & \multicolumn{8}{c}{Unsupervised Representation Learning}                   & \multicolumn{5}{c}{End-to-end Forecasting} \\
\cmidrule{3-14}\multicolumn{2}{c}{} & \multicolumn{2}{c}{Ours} & \multicolumn{2}{c}{TS2Vec} & \multicolumn{2}{c}{TNC} & \multicolumn{2}{c}{CoST} & \multicolumn{2}{c}{Informer} & \multicolumn{2}{c}{TCN}  \\
\midrule
\multicolumn{2}{r}{ L} & MSE   & MAE   & MSE   & MAE   & MSE   & MAE   & MSE   & MAE   & MSE   & MAE   & MSE   & MAE  \\
\midrule
\multicolumn{1}{c|}{\multirow{5}[2]{*}{\begin{sideways}ETTh1\end{sideways}}} & 
%ETTh1 24
\multicolumn{1}{c|}{24} & \textbf{0.377} & \textbf{0.422} & 0.590 & 0.531 & 0.708 & 0.592 & \underline{0.386} & \underline{0.429} & \multicolumn{1}{|c}{0.577} & 0.549 & \multicolumn{1}{r}{0.583} & \multicolumn{1}{r}{0.547} \\
%ETTh1 48
\multicolumn{1}{c|}{} & \multicolumn{1}{c|}{48} & \textbf{0.427} & \textbf{0.454} & 0.624 & 0.555 & 0.749 & 0.619 & \underline{0.437} & \underline{0.464}  &\multicolumn{1}{|c}{0.685} & 0.625 & \multicolumn{1}{r}{0.670} & \multicolumn{1}{r}{0.606} \\
%ETTh1 168
\multicolumn{1}{c|}{} & \multicolumn{1}{c|}{168} & \textbf{0.638} & \textbf{0.577} & 0.762 & 0.639 & 0.884 & 0.699 & \underline{0.643} & \underline{0.582} & \multicolumn{1}{|c}{0.931} & 0.752 & \multicolumn{1}{r}{0.811} & \multicolumn{1}{r}{0.680} \\
%ETTh1 336
\multicolumn{1}{c|}{} & \multicolumn{1}{c|}{336} & \underline{0.815} & \textbf{0.685} & 0.931 & 0.728 & 1.020 & 0.768 & \textbf{0.812} & \underline{0.679} & \multicolumn{1}{|c}{1.128} & 0.873 & \multicolumn{1}{r}{1.132} & \multicolumn{1}{r}{0.815} \\
%ETTh1 720
\multicolumn{1}{c|}{} & \multicolumn{1}{c|}{720} & \textbf{0.956} & \textbf{0.771} & 1.063 & 0.799 & 1.157 & 0.830 & \underline{0.970} & \underline{0.771} &\multicolumn{1}{|c}{1.215} & 1.869 & \multicolumn{1}{r}{1.165} & \multicolumn{1}{r}{0.813} \\
\midrule
%ETTh2 24
\multicolumn{1}{c|}{\multirow{5}[2]{*}{\begin{sideways}ETTh2\end{sideways}}} & \multicolumn{1}{c|}{24} & \textbf{0.336} & \textbf{0.434} & \underline{0.424} & \underline{0.489} & 0.612 & 0.595 & 0.447 & 0.502 & \multicolumn{1}{|c} {0.720} & 0.665 & \multicolumn{1}{r}{0.935} & \multicolumn{1}{r}{0.754} \\
%ETTh2 48
\multicolumn{1}{c|}{} & \multicolumn{1}{c|}{48} & \textbf{0.564} & \textbf{0.571} & \underline{0.619} & \underline{0.605} & 0.840 & 0.716 & 0.699 & 0.637 & \multicolumn{1}{|c}{1.457} & 1.001 & \multicolumn{1}{r}{1.300} & \multicolumn{1}{r}{0.911} \\
%ETTh2 168
\multicolumn{1}{c|}{} & \multicolumn{1}{c|}{168} & \textbf{1.407} & \textbf{0.926} & 1.845 & 1.074 & 2.359 & 1.213 & \underline{1.549} & \underline{0.982} & \multicolumn{1}{|c}{3.489} & 1.515 & \multicolumn{1}{r}{4.017} & \multicolumn{1}{r}{1.579} \\
%ETTh2 336
\multicolumn{1}{c|}{} & \multicolumn{1}{c|}{336} & \textbf{1.640} & \textbf{0.996} & 2.194 & 1.197 & 2.782 & 1.349 & \underline{1.749} & \underline{1.042} & \multicolumn{1}{|c}{2.723} & 1.340 & \multicolumn{1}{r}{3.460} & \multicolumn{1}{r}{1.456} \\
%ETTh2 720
\multicolumn{1}{c|}{} & \multicolumn{1}{c|}{720} & \textbf{1.878} & \textbf{1.065} & 2.636 & 1.370 & 2.753 & 1.394 & \underline{1.971} & \underline{1.092} & \multicolumn{1}{|c}{3.467} & 1.473 & \multicolumn{1}{r}{3.106} & \multicolumn{1}{r}{1.381} \\
\midrule
%ETTm1 24
\multicolumn{1}{c|}{\multirow{5}[2]{*}{\begin{sideways}ETTm1\end{sideways}}} & \multicolumn{1}{c|}{24} & \textbf{0.232} & \textbf{0.314} & 0.453 & 0.444 & 0.522 & 0.472 & \underline{0.246} & \underline{0.329} &\multicolumn{1}{|c}{0.323} & 0.369 & \multicolumn{1}{r}{0.522} & \multicolumn{1}{r}{0.472} \\
%ETTm1 48
\multicolumn{1}{c|}{} & \multicolumn{1}{c|}{48} & \textbf{0.311} & \textbf{0.368} & 0.592 & 0.521 & 0.695 & 0.567 & \underline{0.381} & \underline{0.386} &\multicolumn{1}{|c}{0.494} & 0.503 & \multicolumn{1}{r}{0.542} & \multicolumn{1}{r}{0.508} \\
%ETTm1 96
\multicolumn{1}{c|}{} & \multicolumn{1}{c|}{96} & \textbf{0.360} & \textbf{0.402} & 0.635 & 0.554 & 0.731 & 0.595 & \underline{0.378} & \underline{0.419} &\multicolumn{1}{|c}{0.678} & 0.614 & \multicolumn{1}{r}{0.666} & \multicolumn{1}{r}{0.578} \\
%ETTm1 288
\multicolumn{1}{c|}{} & \multicolumn{1}{c|}{288} & \textbf{0.450} & \textbf{0.467} & 0.693 & 0.597 & 0.818 & 0.649 & \underline{0.472} & \underline{0.486} &\multicolumn{1}{|c}{1.056} & 0.786 & \multicolumn{1}{r}{0.991} & \multicolumn{1}{r}{0.735} \\
%ETTm1 672
\multicolumn{1}{c|}{} & \multicolumn{1}{c|}{672} & \textbf{0.612} & \textbf{0.563} & 0.782 & 0.653 & 0.932 & 0.712 & \underline{0.620} & \underline{0.574} &\multicolumn{1}{|c}{1.192} & 0.926 & \multicolumn{1}{r}{1.032} & \multicolumn{1}{r}{0.756} \\
%ETTm2 24
\midrule
\multicolumn{1}{c|}{\multirow{5}[2]{*}{\begin{sideways}ETTm2\end{sideways}}} & \multicolumn{1}{c|}{24} & \textbf{0.108} & \textbf{0.223} & 0.180 & 0.293 & 0.185 & 0.297 & \underline{0.122} & \underline{0.244} &\multicolumn{1}{|c}{0.173} & 0.301 & \multicolumn{1}{r}{0.180} & \multicolumn{1}{r}{0.324} \\
%ETTm2 48
\multicolumn{1}{c|}{} & \multicolumn{1}{c|}{48} & \textbf{0.164} & \textbf{0.285} & 0.244 & 0.350 & 0.264 & 0.360 & \underline{0.183} & \underline{0.305} & \multicolumn{1}{|c}{0.303} & 0.409 & \multicolumn{1}{r}{0.204} & \multicolumn{1}{r}{0.327} \\
%ETTm2 96
\multicolumn{1}{c|}{} & \multicolumn{1}{c|}{96} & \textbf{0.271} & \textbf{0.376} & 0.360 & 0.427 & 0.389 & 0.458 & \underline{0.294} & \underline{0.394} & \multicolumn{1}{|c}{0.365} & 0.453 & \multicolumn{1}{r}{3.041} & \multicolumn{1}{r}{1.330} \\
%ETTm2 288
\multicolumn{1}{c|}{} & \multicolumn{1}{c|}{288} & \textbf{0.716} & \textbf{0.646} & \underline{0.723} & \underline{0.639} & 0.920 & 0.788 & \underline{0.723} & 0.652 & \multicolumn{1}{|c}{1.047}& 0.804 & \multicolumn{1}{r}{3.162} & \multicolumn{1}{r}{1.337} \\
%ETTm2 672
\multicolumn{1}{c|}{} & \multicolumn{1}{c|}{672} & \textbf{1.600} & \textbf{0.979} & \underline{1.753} & \underline{1.007} & 2.164 & 1.135 & 1.899 & 1.073 & \multicolumn{1}{|c}{3.126} & 1.302 & \multicolumn{1}{r}{3.624} & \multicolumn{1}{r}{1.484} \\
\midrule
%exchange 24
\multicolumn{1}{c|}{\multirow{5}[2]{*}{\begin{sideways}Exchange\end{sideways}}} & \multicolumn{1}{c|}{24} & \textbf{0.059} & \textbf{0.172} & 0.108 & 0.252 & \underline{0.105} & \underline{0.236} & 0.136 & 0.291 &\multicolumn{1}{|c}{ 0.611} & 0.626 & \multicolumn{1}{r}{2.483} & \multicolumn{1}{r}{1.327} \\
%exchange 48
\multicolumn{1}{c|}{} & \multicolumn{1}{c|}{48} & \textbf{0.135} & \textbf{0.265} & 0.200 & 0.341 & \underline{0.162} & \underline{0.270} & 0.250 & 0.387 &\multicolumn{1}{|c}{0.680} & 0.644 & \multicolumn{1}{r}{2.328} & \multicolumn{1}{r}{1.256} \\
%exchange 168
\multicolumn{1}{c|}{} & \multicolumn{1}{c|}{168} & 0.713 & 0.635 & \underline{0.412} & \underline{0.492} & \textbf{0.397} & \textbf{0.480} & 0.924 & 0.762 &\multicolumn{1}{|c}{1.097} & 0.825 & \multicolumn{1}{r}{2.372} & \multicolumn{1}{r}{1.279} \\
%exchange 336
\multicolumn{1}{c|}{} & \multicolumn{1}{c|}{336} & 1.409 & 0.938 & \underline{1.339} & \underline{0.901} & \textbf{1.008} & \textbf{0.866} & 1.774 & 1.063 &\multicolumn{1}{|c}{1.672} & 1.036 & \multicolumn{1}{r}{3.113} & \multicolumn{1}{r}{1.459} \\
%exchange 720
\multicolumn{1}{c|}{} & \multicolumn{1}{c|}{720} & \textbf{1.628} & \textbf{1.056} & 2.114 & 1.125 & \underline{1.989} & \underline{1.063} & 2.160 & 1.209 & \multicolumn{1}{|c}{2.478} & 1.310 & \multicolumn{1}{r}{3.150} & \multicolumn{1}{r}{1.458} \\
\midrule
%WTH 24
\multicolumn{1}{c|}{\multirow{5}[2]{*}{\begin{sideways}Weather\end{sideways}}} & \multicolumn{1}{c|}{24} & \textbf{0.298} & \textbf{0.359} & \underline{0.308} & 0.364 & 0.320 & 0.373 & \textbf{0.298} & \underline{0.360} &\multicolumn{1}{|c}{0.335} & 0.381 & \multicolumn{1}{r}{0.321} & \multicolumn{1}{r}{0.367} \\
%WTH 48
\multicolumn{1}{c|}{} & \multicolumn{1}{c|}{48} & \textbf{0.359} & \textbf{0.410} & \underline{0.375} & 0.417 & 0.380 & 0.421 & \textbf{0.359} & \underline{0.411} &\multicolumn{1}{|c}{0.395} & 0.459 & \multicolumn{1}{r}{0.386} & \multicolumn{1}{r}{0.423} \\
%WTH 168
\multicolumn{1}{c|}{} & \multicolumn{1}{c|}{168} & \textbf{0.426} & \textbf{0.461} & 0.496 & 0.506 & 0.479 & 0.495 & \underline{0.464} & \underline{0.491} &\multicolumn{1}{|c}{0.608} & 0.567 & \multicolumn{1}{r}{0.491} & \multicolumn{1}{r}{0.501} \\
%WTH 336
\multicolumn{1}{c|}{} & \multicolumn{1}{c|}{336} & {0.504} & {0.520} & 0.532 & 0.533 & 0.505 & 0.514 & \textbf{0.497} & \textbf{0.517} &\multicolumn{1}{|c}{0.702} & 0.620 & \multicolumn{1}{r}{\underline{0.502}} & \multicolumn{1}{r}{\underline{0.507}} \\
%WTH 720
\multicolumn{1}{c|}{} & \multicolumn{1}{c|}{720} & {0.535} & \underline{0.542} & 0.567 & 0.558 & 0.543 & 0.547 & \underline{0.533} & \underline{0.542} & \multicolumn{1}{|c}{0.831} & 0.731 & \multicolumn{1}{r} {\textbf{0.498}} & \multicolumn{1}{r}{\textbf{0.508}} \\
\midrule
\multicolumn{2}{c}{Avg.} & \textbf{0.664} & \textbf{0.562} & 0.818 & 0.632 & 0.909 & 0.669 & \underline{0.746} & \underline{0.603} & 1.151 & 0.812 & 1.560 & 0.884  \\
\bottomrule
\multicolumn{14}{l}{- The results of TS2Vec and CoST on ETTm2, Exchange, and Weather datasets are implemented by us.}
\end{tabular}
    }
    \caption{Multivariate forecasting results. The best results are highlighted in bold, and the second-best results are highlighted with an underline. $L$ denotes the predicted horizons of datasets.%The predicted horizon is $L\in \{24,48,168,336,720\}$ for ETTh1, ETTh2, Exchange and Weather, whereas it is $L \in \{24,48,96,288,672\}$ for ETTm1 and ETTm2. \
    The performance is measured in mean-squared error (MSE) and mean-absolute error (MAE).}
    \label{tab:main-multi}%
\end{table*}%

\subsection{Experimental setup}
We divide all datasets into training, validation, and test sets in the ratio of 6:2:2. Throughout the evaluation stage, the model parameters are frozen to output representations. 

The input time series are projected to a 64-dimensional latent space using a convolutional projector. The multi-scale convolutions further encode the projected vectors into a 320-dimensional latent space (i.e., $C'$ = 320). We cut the original time series into sub-sequences of length T = 402, where each sub-sequence serves as a training sample. Within each sample, the first 201 timestamps correspond to its \textit{history} view and the subsequent 201 timestamps to its \textit{future} view. 
The cosine similarity loss is optimized using stochastic gradient descent (SGD) optimizer with a learning rate of 0.001, a momentum of 0.9, and a weight decay of 0.0001. We trained 500 epochs for all datasets with a batch size of 8. 

We set the predicted horizons $L \in \{24,48,168,336,720\}$ for dataset ETTh1, ETTh2, Exchange, and Weather. For dataset ETTm1 and ETTm2, we set $L \in \{24,48,96,288,672\}$. We select the best ridge regression model using the validation set and then use it to report the forecasting error on the test set. Mean-squared-error (MSE) and mean-absolute-error (MAE) are used to evaluate our results. More details about the experimental setup and training process are included in the appendix, and codes for reproducing the results will be available upon acceptance.

\subsection{Results}
Table~\ref{tab:main-multi} summarizes the average results of multivariate forecasting with five runs. Overall, our model, SimTS, outperforms all the
representation learning baselines in the multivariate setting on most of the datasets by a large margin. When looking at the average performance across six datasets, SimTS outperforms TS2Vec by 18.8\% (MSE) and 11.1\%(MAE), TNC by 36.9\% (MSE) and 16.0\%(MAE), and CoST by 11.0\% (MSE) and 6.8\%(MAE). Additionally, when examining the performance on each dataset individually, SimTS outperforms TS2Vec on all six datasets and outperforms TNC and CoST on five out of six datasets while performing comparably or slightly worse on one of the six datasets. We believe one probable explanation is that the Exchange dataset is less stationary, and the pattern of data adjacent in time (i.e., in a neighborhood) can be discriminated from the pattern of data far away. Such neighborhood patterns can be found via TNC, which leads to better performance. On the other hand, the weather dataset is more stationary, which means CoST can use season-trend disentanglement to extract useful information and thus achieves better performance. 

Although CoST and TNC perform better in some datasets, SimTS achieves overall state-of-the-art performance across all datasets. This suggests that our approach is general and robust across a wide range of time series datasets.

%---------------------------------------------------

\section{Ablation Study}
In this section, we present a systematic ablation study to examine the different components and assumptions in our model. We also investigate assumptions in the baseline models to assess their influence on forecasting performance.
\subsection{Backbones}
First, we examine the importance of our encoder network structure design. To test the contribution of the convolutional network structure as our encoding network, we substitute the convolutional layers with the TCN~\cite{CNN} and LSTM~\cite{lstm} networks with comparable parameter sizes. Table~\ref{tab:backbone} shows the forecasting results on ETT datasets. In both univariate and multivariate forecasting, the convolutional layer in our model performs better than TCN and LSTM, demonstrating the efficiency of our encoder for encoding time series representations.

\begin{table}[H]
  \small
  \centering
  \setlength{\tabcolsep}{5pt} 
  {
    \begin{tabular}{l|rrrrrr}
    \toprule
    Backbones & \multicolumn{2}{c}{Ours}       & \multicolumn{2}{c}{TCN}      & \multicolumn{2}{c}{LSTM} \\
    %\midrule
    %Methods & \multicolumn{2}{c}{TS2Vec} & \multicolumn{2}{c}{CoST} & \multicolumn{2}{c}{TS2Vec} & \multicolumn{2}{c}{CoST} & \multicolumn{2}{c}{TS2Vec} & \multicolumn{2}{c}{CoST} \\
    \midrule
          & \multicolumn{1}{l}{MSE} & \multicolumn{1}{l}{MAE} & \multicolumn{1}{l}{MSE} & \multicolumn{1}{l}{MAE} & \multicolumn{1}{l}{MSE} & \multicolumn{1}{l}{MAE}  \\
    \midrule
    Multivariate & \textbf{0.688} & \textbf{0.601} & 0.912 & 0.674 & 2.124 & 0.827  \\
    Univariate & \textbf{0.041} & \textbf{0.220} & 0.134 & 0.250 & 1.750 & 1.274  \\
    \bottomrule
    \end{tabular}
    }
  
\caption{Ablation study of different backbone architectures on ETT datasets.} 
\label{tab:backbone}%
\end{table}%

\subsection{Negative Samples}

\begin{table*}[ht]
  \small
  \centering
  \setlength{\tabcolsep}{5pt} %\setlength{\tabcolsep}{5pt}
  {
    \begin{tabular}{lrrrrrrrrrrrr}
    \toprule
    Datasets & \multicolumn{2}{c}{ETTh1}       & \multicolumn{2}{c}{ETTh2}      & \multicolumn{2}{c}{ETTm1} &\multicolumn{2}{c}{ETTm2} &\multicolumn{2}{c}{Exchange} &\multicolumn{2}{c}{Weather} \\
    \midrule
          & \multicolumn{1}{l}{MSE} & \multicolumn{1}{l}{MAE} & \multicolumn{1}{l}{MSE} & \multicolumn{1}{l}{MAE} & \multicolumn{1}{l}{MSE} & \multicolumn{1}{l}{MAE} &\multicolumn{1}{l}{MSE} & \multicolumn{1}{l}{MAE}& \multicolumn{1}{l}{MSE} & \multicolumn{1}{l}{MAE} &\multicolumn{1}{l}{MSE} & \multicolumn{1}{l}{MAE}  \\
    
    \midrule
    SimTS & {0.642} & {0.582} & 1.165 & 0.798 & 0.393 & 0.423 & 0.572 & 0.502 &0.789 &0.613 &0.424 &0.458\\
    SimTS w/ neg$^\ddag$ & {0.685} & {0.632} & 1.544 & 0.938 & 0.392 & 0.441 &0.747 &0.572&1.405&0.769&0.434&0.468 \\
    \bottomrule
    \multicolumn{12}{l}{$^\ddag$ With negative samples and InfoNCE loss in Equation~\ref{eqn:infoNCE}.}
    \end{tabular}%
    }
  \caption{Ablation study of negative samples on multivariate forecasting across ETT datasets.}
  \label{tab:neg}
\end{table*}

Negative pairs, if not constructed carefully, could depreciate the model performance in terms of representation power.  TS2Vec~\cite{yue2022ts2vec} and CoST~\cite{woo2022cost} use sub-sequences of other instances or various timestamps as the negative pairs for contrastive learning. However, our model SimTS outperforms them in the absence of negative pairs, implying that the selection of negative pairs in CoST and TS2Vec may be inaccurate and result in sub-optimal performance. To further demonstrate the influence of negative samples, we construct negative pairs by following SimCLR~\cite{simclr} to test our model result with and without negative pairs. We replace the cosine similarity loss in Equation~\eqref{eqn:loss} with the loss that is used in~\citet{simclr} and~\citet{cpc} to consider the negative pairs together with the positive pairs. Given a mini-batch $\mathcal{D} = \{X_1,X_2,...,X_N\}$ of $N$ samples with length $T$ and its encoded representations \{$Z_1,Z_2,...,Z_N$\}, we optimize: 
\begin{equation}
\label{eqn:infoNCE}
\mathcal L_{\theta, \phi}^{\text{NCE}}(\mathcal{D}) = -\frac{1}{T}\sum_{t=1}^{T}\left[\log\frac{\exp (\hat{z}^f_t\cdot z^f_{t,+})}{\sum_{i=1}^{N} \exp (\hat{z}^f_t\cdot z^f_{t,i})}\right],
\end{equation} where $\hat{z}^f_{t,+}= G_\theta(z^h_t)$, and $z^f_{t,i}$ denotes latent representation of the $t$-th timestamp of the $i$-th sample from $\mathcal{D}$. The numerator calculates the similarity between predicted and encoded future representations, which is the positive pair in our framework. The denominator calculates the similarities between the negative pairs, which are the predicted future representation and encoded representations from other samples within $\mathcal{D}$. Table~\ref{tab:neg} shows the forecasting results with and without including the negative samples. In particular, it demonstrates that negative samples generally decrease performance in most of the datasets we tested. These results confirm that adding negative pairs to our proposed method leads to suboptimal performance. However, this does not mean that including negative pairs overall is not useful; it simply implies that the current approaches to constructing negative pairs are inefficient. Thus, future research should be dedicated to coming up with better ways to construct negative pairs.

\subsection{Stop-Gradient Operation}

\begin{figure*}
     \centering
     \begin{subfigure}[b]{0.25\textwidth}
         \centering
         \includegraphics[width=0.9\textwidth]{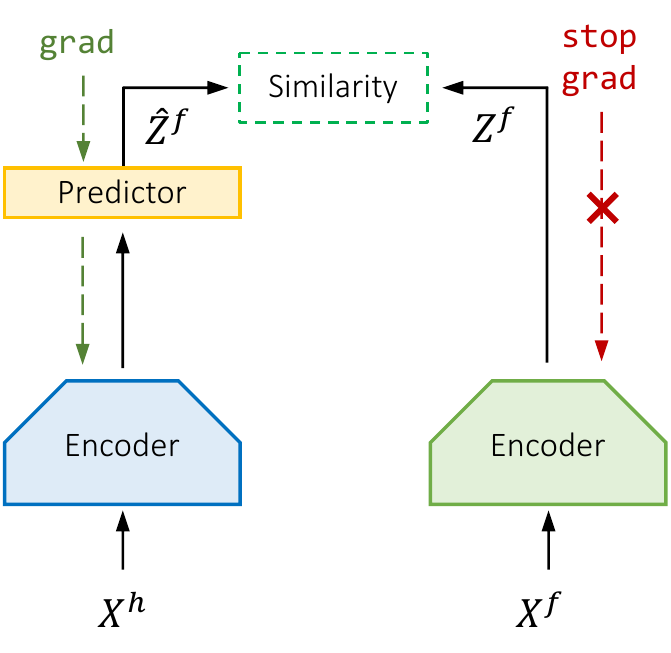}
         \caption{SimTS}
         \label{fig:SimTS}
     \end{subfigure}
     \hfill
     \begin{subfigure}[b]{0.25\textwidth}
         \centering
         \includegraphics[width=0.9\textwidth]{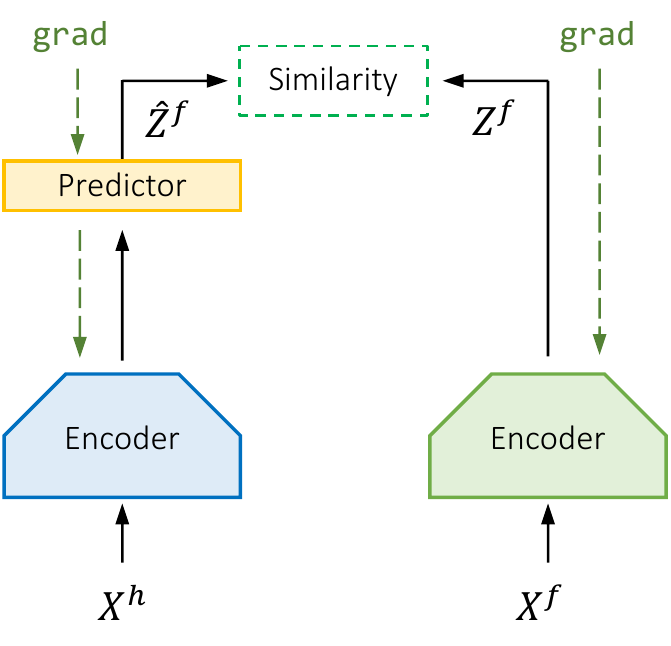}
         \caption{SimTS w/o SG}
         \label{fig:nogradient}
     \end{subfigure}
     \hfill
     \begin{subfigure}[b]{0.25\textwidth}
         \centering
         \includegraphics[width=0.95\textwidth]{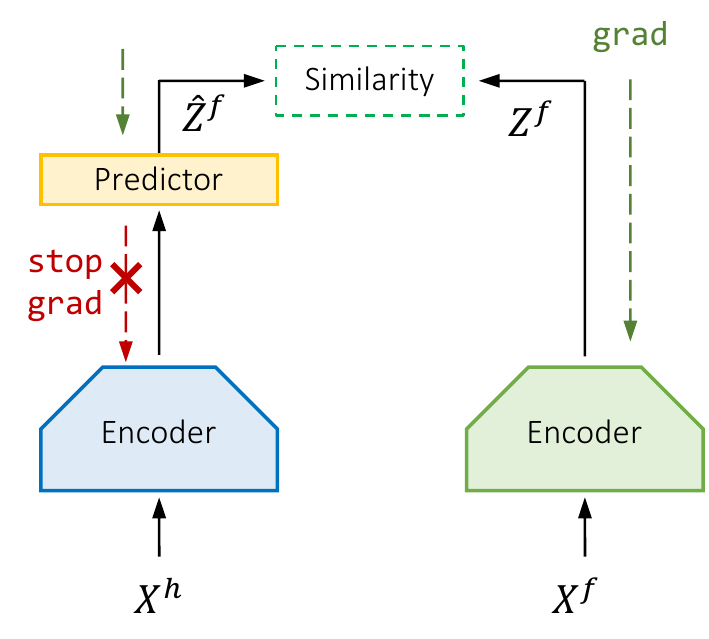}
         \caption{RevSimTS}
         \label{fig:rev}
     \end{subfigure}
        \caption{Ablation study of stop-gradient operation. (a) SimTS architecture. (b) SimTS without stop-gradient operation. (c) RevSimTS with stop-gradient on the \textit{history} encoding path. }
        \label{fig:three graphs}
\end{figure*}

In SimTS, we apply a stop-gradient operation on the \textit{future} encoding path during the optimization. To test the effect of this operation on the overall model performance, we did an ablation study on what happens if we remove this operation, or apply it on the history encoding path instead of the \textit{future} encoding path, see Figure~\ref{fig:three graphs}. When we apply the stop gradient on the history encoding path, as it is shown in Figure~\ref{fig:rev}, the model optimizes the loss by pushing \textit{future} representations $Z^f$ towards the \textit{future} predictions $\hat{Z}^f$. We refer to this model as RevSimTS (SimTS with reverse stop-gradient, Figure~\ref{fig:rev}). As shown in Table~\ref{tab:gradient}, we observe that either the removal of the stop-gradient on the \textit{future} encoding path (Figure~\ref{fig:nogradient}) or moving the stop-gradient to the history encoding path causes a significant decrease in performance, supporting our argument that the stop-gradient operation in the future encoding path leads to optimal performance.

\begin{table}[H]
  \small
  \centering
  \setlength{\tabcolsep}{5pt} %\setlength{\tabcolsep}{5pt}
  {
    \begin{tabular}{l|cccccc}
    \toprule
    Model & \multicolumn{2}{c}{SimTS}       & \multicolumn{2}{c}{SimTS w/o SG$^\dag$}      & \multicolumn{2}{c}{RevSimTS} \\
    %\midrule
    %Methods & \multicolumn{2}{c}{TS2Vec} & \multicolumn{2}{c}{CoST} & \multicolumn{2}{c}{TS2Vec} & \multicolumn{2}{c}{CoST} & \multicolumn{2}{c}{TS2Vec} & \multicolumn{2}{c}{CoST} \\
    \midrule
          & \multicolumn{1}{l}{MSE} & \multicolumn{1}{l}{MAE} & \multicolumn{1}{l}{MSE} & \multicolumn{1}{l}{MAE} & \multicolumn{1}{l}{MSE} & \multicolumn{1}{l}{MAE}  \\
    \midrule
    ETTh1 & \textbf{0.642} & \textbf{0.582} & 0.783 &0.663 & 0.762 & 0.634  \\
    ETTh2 & \textbf{1.165} & \textbf{0.798} & 2.940 & 1.490 & 3.128 & 1.449  \\
    ETTm1 & \textbf{0.393} & \textbf{0.432} & 0.681 & 0.609 & 0.551 & 0.525  \\
    ETTm2 & \textbf{0.572} & \textbf{0.502} & 1.315 & 0.863 & 1.186 & 0.796  \\
    Exchange & \textbf{0.789} & \textbf{0.613} & 1.808 & 1.062 & 1.398 & 0.900  \\
    Weather & \textbf{0.424} & \textbf{0.458} & 0.605 & 0.592 & 0.485 & 0.512  \\
    \bottomrule
    \multicolumn{5}{l}{$^\dag$ SimTS without stop-gradient operation} \\
    \end{tabular}
    }
  \caption{Ablation study of stop-gradient operation on multivariate forecasting across ETT datasets.}
  \label{tab:gradient}%
\end{table}
\subsection{Disentanglement Assumption}
\label{ab:season}

\begin{table}[htbp]
    \small
    \centering
    \setlength{\tabcolsep}{5pt}
    \begin{tabular}{l|cccc} \toprule
    Datasets & Exchange & ETTm2 & ETTm1 &  Weather\\\midrule
     ADF Test Stat. &  -1.889 & -6.225 & -14.985 & -26.661  \\
    \midrule
     CoST  & 0.975 & 0.822  &0.492 & 0.439 \\
     CoST w/o SD$^\dag$& 0.899 & 0.754 & 0.466 & 0.440 \\
     CoST w/o aug.$^\ddag$&0.865 &0.986  &0.493  & 0.462 \\
     CoST w/ mask$^\S$&1.223 &0.664  &1.041  & 0.502   \\
     \midrule
     Diff. w/o SD & 0.076 \textcolor{green}{$\uparrow$}& 0.068  \textcolor{green}{$\uparrow$}&0.026  \textcolor{green}{$\uparrow$}&0.001 \textcolor{red}{$\downarrow$}\\
     Diff. w/o aug. &0.119  \textcolor{green}{$\uparrow$}&0.164 \textcolor{red}{$\downarrow$}&0.001 \textcolor{red}{$\downarrow$}&0.023 \textcolor{green}{$\uparrow$}\\
     Diff. w/ mask &0.248 \textcolor{red}{$\downarrow$}&0.158 \textcolor{green}{$\uparrow$}&0.549 \textcolor{red}{$\downarrow$}&0.063 \textcolor{red}{$\downarrow$}\\
     \bottomrule
     \multicolumn{5}{l}{$^\dag$ Seasonal disentanglement} \\
     \multicolumn{5}{l}{$^\ddag$ Discard the augmentations proposed in~\cite{woo2022cost}}\\
     \multicolumn{5}{l}{$^\S$ Timestamp masking proposed in~\cite{yue2022ts2vec}}\\
     \multicolumn{5}{l}{\textcolor{green}{$\uparrow$}/\textcolor{red}{$\downarrow$} indicates performance increase/decrease}
    \end{tabular}
\caption{The average multivariate forecasting results from changing the season-trend disentanglement and data augmentation modules in CoST.}
\label{tab:disentangle}
\end{table}
To demonstrate that the season-trend disentanglement as proposed in CoST~\cite{woo2022cost} may not work well for different types of datasets, especially on the less stationary data, we conduct an ablation study by removing the season disentanglement in CoST. The original season-trend disentanglement is performed by applying Fourier transform to the data and using an affine transformation to extract feature correlations in the frequency domain. We substitute this process by performing the same affine transformation on the original data without applying Fourier transform. The results are shown in Table~\ref{tab:disentangle}, where ``w/o SD" denotes CoST without the seasonal disentanglement. Besides, we adopt the Augmented Dick-Fuller (ADF) test statistic~\cite{adf} proposed in~\cite{liu2022nonstationary} to measure the degree of stationarity. A smaller ADF score indicates higher stationarity. We observe that seasonal disentanglement can improve the forecasting outcomes for the Weather dataset, which exhibits significant stationarity. However, the seasonal disentanglement impairs predicting ability in less stationary datasets like Exchange and ETTm2, supporting our claim that the seasonal disentanglement assumption is misleading in some datasets and lacks generality. 

\subsection{
Data Augmentation for Constructing Views}

Data augmentation is a common method to generate positive pairs in contrastive learning. However, current augmentation methods for time series may impair the performance of forecasting. We conduct ablation studies to demonstrate the influences of data augmentations. CoST uses three types of data augmentation:  scaling, shifting, and jittering. On the other hand, TS2Vec randomly masks timestamps in a sample to construct views. Therefore, we implement two ablation experiments for CoST: (1) eliminating data augmentation and (2) adding random masks. Table~\ref{tab:disentangle} shows the results of the two experiments, where ``w/o aug" denotes CoST without its original augmentation methods and ``w/mask" denotes CoST using random masks as augmentation. Our experiments show that the original data augmentation in CoST can potentially result in lower performances, and adding random masks impairs performances for most datasets. These findings do not imply that data augmentation is not effective in general; rather, they demonstrate that finding efficient augmentation techniques applicable to various time series is challenging, and better methods for augmenting time series data need to be developed.
\section{Conclusion}

This paper proposes SimTS, a simple representation learning framework based on contrastive learning that does not require negative pairs. We conducted an extensive study to test our proposed model and compared it to other existing representation learning models for time series forecasting. Our general aim was to challenge the assumptions and components that are widely used in these models. Our study reveals that current representation learning methods are not universally applicable to different types of time series data. Some of the components used in these models might be unnecessary and can even negatively impact performance in some cases. This means that existing models based on contrastive learning for time series forecasting are highly dependent on the specific dataset being used, and careful consideration is necessary when deploying them.

Our proposed model, however, addresses some of the limitations by providing a simplified and robust contrastive learning model achieving better performance across different datasets compared to state-of-the-art methods. Moving forward, we plan to extend our framework to handle more challenging data such as irregular time series and explore efficient data augmentation methods for time series forecasting.

% \clearpage 
\section{Acknowledgements}
This work is supported by the Swiss National Science Foundation (project 201184).

\newpage
\bibliography{paper}
\bibliographystyle{icml2023}

%%%%%%%%%%%%%%%%%%%%%%%%%%%%%%%%%%%%%%%%%%%%%%%%%%%%%%%%%%%%%%%%%%%%%%%%%%%%%%%
%%%%%%%%%%%%%%%%%%%%%%%%%%%%%%%%%%%%%%%%%%%%%%%%%%%%%%%%%%%%%%%%%%%%%%%%%%%%%%%
% APPENDIX
%%%%%%%%%%%%%%%%%%%%%%%%%%%%%%%%%%%%%%%%%%%%%%%%%%%%%%%%%%%%%%%%%%%%%%%%%%%%%%%
%%%%%%%%%%%%%%%%%%%%%%%%%%%%%%%%%%%%%%%%%%%%%%%%%%%%%%%%%%%%%%%%%%%%%%%%%%%%%%%
\newpage
\phantomsection
\appendix
\onecolumn
\section{Summary of Time Series Forecasting Datasets}
\begin{table}[!ht]
  \small
  \centering
  \setlength{\tabcolsep}{5pt} 
  {
    \begin{small}
    \begin{threeparttable}
    \setlength{\tabcolsep}{8.5pt}
    \begin{tabular}{lcccc}
    \toprule
    \multicolumn{1}{l}{Dataset} & \multicolumn{1}{l}{Variable Number} & \multicolumn{1}{l}{Sampling Frequency} & \multicolumn{1}{l}{Total Observations} & \multicolumn{1}{l}{ADF Test Statistic} \\ \toprule
    ETTh1& 7&\multicolumn{1}{c}{1 Hour} &\multicolumn{1}{c}{17,420}& \multicolumn{1}{c}{-5.909}\\
    ETTh2&7&\multicolumn{1}{c}{1 Hour} &\multicolumn{1}{c}{17,420}& \multicolumn{1}{c}{-4.136}\\
    ETTm1&7&\multicolumn{1}{c}{15 Minutes} &\multicolumn{1}{c}{69,680}& \multicolumn{1}{c}{-14.985}\\
    ETTm2                       & 7                                      & \multicolumn{1}{c}{15 Minutes}         & \multicolumn{1}{c}{69,680}            & \multicolumn{1}{c}{-6.225}            \\
    Exchange                    & 8                                      & \multicolumn{1}{c}{1 Day}              & \multicolumn{1}{c}{7,588}             & \multicolumn{1}{c}{-1.889}             \\
    %ILI                     & 7                         & \multicolumn{1}{c}{1 Week}             & \multicolumn{1}{c}{966}               & \multicolumn{1}{c}{-5.406}             \\
    
    %Electricity                 & 321                                    & \multicolumn{1}{c}{1 Hour}             & \multicolumn{1}{c}{26,304}            & \multicolumn{1}{c}{-8.483}             \\
    %Traffic                     & 862                                    & \multicolumn{1}{c}{1 Hour}             & \multicolumn{1}{c}{17,544}            & \multicolumn{1}{c}{-15.046}            \\
    Weather                     & 21                                     & \multicolumn{1}{c}{10 Minutes}         & \multicolumn{1}{c}{52,695}            & \multicolumn{1}{c}{-26.661}            \\ \bottomrule
    \end{tabular}
%   \begin{tablenotes}
%         \footnotesize
% \item[*] A smaller ADF test statistic indicates a more stationary dataset.
%   \end{tablenotes}
  \end{threeparttable}
  \end{small}
    }
  \caption{Summary of datasets. Smaller ADF test statistic indicates a more stationary dataset.} 
\label{tab:dataset}
\end{table}
\section{Details on baselines}
The seven baselines' descriptions and implementations are listed below. We reproduce the results of CoST, TS2Vec, TNC, and Informer for dataset ETTm2, Exchange and Weather. Other results are taken from~\citet{woo2022cost} and~\citet{wang2022last}. Unless otherwise stated, we employ the parameters specified in the respective papers.

\textbf{CoST~\cite{woo2022cost}}: CoST performs season-trend disentanglement to learn seasonal and trend representations separately by using Fourier Transform. The final representation for forecasting is the concatenation of the seasonal and trend representation. We run their code from \url{https://github.com/salesforce/CoST}.

\textbf{TS2Vec~\cite{yue2022ts2vec}}: TS2Vec designs a hierarchical contrastive learning framework to learn a universal time series representation. It employs timestep masks as the data augmentation and temporal convolutions to encode the latent representations. We reproduce their experiments from their publicly available code: \url{https://github.com/yuezhihan/ts2vec}

\textbf{TNC~\cite{tonekaboni2021tnc}}: TNC is an unsupervised representation learning method that makes sure the latent representations from a neighborhood are distinguishable from representations outsides the neighborhood. We use their open source code from \url{https://github.com/sanatonek/TNC_representation_learning}. Following the setup in TS2Vec, we use the casual TCN encoder proposed in TS2Vec to replace the original encoder in TNC. 

\textbf{Informer~\cite{informer}}: Informer is designed based on the transformer for long sequence time series forecasting. It consists of three major components: a ProbSparse self-attention mechanism, a self-attention distilling mechanism, and a generative style decoder. We use their code from \url{https://github.com/zhouhaoyi/Informer2020}

\textbf{TCN~\cite{wavenet}}: TCN proposes dilated convolutions for time series data. A stack of ten residual blocks with a hidden size of 64 is added to the encoder in TS2Vec. Their public source code can be achieved at \url{https://github.com/locuslab/TCN}.  
\section{Results for Univariate Setting}
\begin{table}[H]
  \small
  \centering
  \setlength{\tabcolsep}{5pt} 
  {
    % Table generated by Excel2LaTeX from sheet 'Main Results'
\begin{tabular}{cccccccccccccc}
\toprule
\multicolumn{1}{c}{\multirow{1}[2]{*}{Methods}} & \multicolumn{8}{c}{Unsupervised Representation Learning}                   & \multicolumn{5}{c}{End-to-end Forecasting} \\
\cmidrule{3-14}\multicolumn{2}{c}{} & \multicolumn{2}{c}{Ours} & \multicolumn{2}{c}{TS2Vec} & \multicolumn{2}{c}{TNC} & \multicolumn{2}{c}{CoST} & \multicolumn{2}{c}{Informer} & \multicolumn{2}{c}{TCN}  \\
\midrule
\multicolumn{2}{r}{ Metrics} & MSE   & MAE   & MSE   & MAE   & MSE   & MAE   & MSE   & MAE   & MSE   & MAE   & MSE   & MAE  \\
\midrule
\multicolumn{1}{c|}{\multirow{5}[2]{*}{\begin{sideways}ETTh1\end{sideways}}} & 
%ETTh1 24
\multicolumn{1}{c|}{24} & \textbf{0.036} & \textbf{0.143} & 0.039 & 0.151 & 0.057 & 0.184 & 0.040 & 0.152 & \multicolumn{1}{|c}{0.098} & 0.147 & \multicolumn{1}{r}{0.104} & \multicolumn{1}{r}{0.254} \\
%ETTh1 48
\multicolumn{1}{c|}{} & \multicolumn{1}{c|}{48} & \textbf{0.054} & \textbf{0.176} & 0.062 & 0.189 & 0.094 & 0.239 & 0.060 & 0.186  &\multicolumn{1}{|c}{0.158} & 0.319 & \multicolumn{1}{r}{0.206} & \multicolumn{1}{r}{0.366} \\
%ETTh1 168
\multicolumn{1}{c|}{} & \multicolumn{1}{c|}{168} & \textbf{0.084} & \textbf{0.216} & 0.142 & 0.291 & 0.171 & 0.329 & 0.097 & 0.236 & \multicolumn{1}{|c}{0.183} & 0.346 & \multicolumn{1}{r}{0.462} & \multicolumn{1}{r}{0.586} \\
%ETTh1 336
\multicolumn{1}{c|}{} & \multicolumn{1}{c|}{336} & \textbf{0.100} & \textbf{0.239} & 0.160 & 0.316 & 0.179 & 0.345 & 0.112 & 0.258 & \multicolumn{1}{|c}{0.222} & 0.387 & \multicolumn{1}{r}{0.422} & \multicolumn{1}{r}{0.564} \\
%ETTh1 720
\multicolumn{1}{c|}{} & \multicolumn{1}{c|}{720} & \textbf{0.126} & \textbf{0.277} & 0.179 & 0.345 & 0.235 & 0.408 & 0.148 & 0.306 &\multicolumn{1}{|c}{0.269} & 0.435 & \multicolumn{1}{r}{0.438} & \multicolumn{1}{r}{0.578} \\
\midrule
%ETTh2 24
\multicolumn{1}{c|}{\multirow{5}[2]{*}{\begin{sideways}ETTh2\end{sideways}}} & \multicolumn{1}{c|}{24} & \textbf{0.077} & \textbf{0.206} & 0.097 & 0.230 & 0.097 & 0.238 & 0.079 & 0.207 & \multicolumn{1}{|c} {0.093} & 0.240 & \multicolumn{1}{r}{0.109} & \multicolumn{1}{r}{0.251} \\
%ETTh2 48
\multicolumn{1}{c|}{} & \multicolumn{1}{c|}{48} & \textbf{0.116} & \textbf{0.259} & 0.124 & 0.274 & 0.131 & 0.281 & 0.118 & 0.259 & \multicolumn{1}{|c}{0.155} & 0.314 & \multicolumn{1}{r}{0.147} & \multicolumn{1}{r}{0.302} \\
%ETTh2 168
\multicolumn{1}{c|}{} & \multicolumn{1}{c|}{168} & 0.191 & 0.340 & 0.198 & 0.355 & 0.197 & 0.354 & \textbf{0.189} & \textbf{0.339} & \multicolumn{1}{|c}{0.232} & 0.389 & \multicolumn{1}{r}{0.209} & \multicolumn{1}{r}{0.366} \\
%ETTh2 336
\multicolumn{1}{c|}{} & \multicolumn{1}{c|}{336} & \textbf{0.199} & \textbf{0.354} & 0.205 & 0.364 & 0.207 & 0.366 & 0.206 & 0.360 & \multicolumn{1}{|c}{0.263} & 0.417 & \multicolumn{1}{r}{0.237} & \multicolumn{1}{r}{0.391} \\
%ETTh2 720
\multicolumn{1}{c|}{} & \multicolumn{1}{c|}{720} & \textbf{0.212} & \textbf{0.370} & 0.208 & 0.371 & 0.207 & 0.370 & 0.214 & 0.371 & \multicolumn{1}{|c}{0.277} & 0.431 & \multicolumn{1}{r}{0.200} & \multicolumn{1}{r}{0.367} \\
\midrule
%ETTm1 24
\multicolumn{1}{c|}{\multirow{5}[2]{*}{\begin{sideways}ETTm1\end{sideways}}} & \multicolumn{1}{c|}{24} & \textbf{0.013} & \textbf{0.084} & 0.016 & 0.093 & 0.019 & 0.103 & 0.015 & 0.088 &\multicolumn{1}{|c}{0.030} & 0.137 & \multicolumn{1}{r}{0.027} & \multicolumn{1}{r}{0.127} \\
%ETTm1 48
\multicolumn{1}{c|}{} & \multicolumn{1}{c|}{48} & \textbf{0.024} & \textbf{0.112} & 0.028 & 0.126 & 0.045 & 0.162 & 0.025 & 0.117 &\multicolumn{1}{|c}{0.069} & 0.203 & \multicolumn{1}{r}{0.040} & \multicolumn{1}{r}{0.154} \\
%ETTm1 96
\multicolumn{1}{c|}{} & \multicolumn{1}{c|}{96} & \textbf{0.041} & \textbf{0.143} & 0.045 & 0.162 & 0.054 & 0.178 & 0.038 & 0.147 &\multicolumn{1}{|c}{0.194} & 0.372 & \multicolumn{1}{r}{0.097} & \multicolumn{1}{r}{0.246} \\
%ETTm1 288
\multicolumn{1}{c|}{} & \multicolumn{1}{c|}{288} & \textbf{0.098} & \textbf{0.207} & 0.095 & 0.235 & 0.142 & 0.290 & 0.077 & 0.209 &\multicolumn{1}{|c}{0.401} & 0.544 & \multicolumn{1}{r}{0.305} & \multicolumn{1}{r}{0.455} \\
%ETTm1 672
\multicolumn{1}{c|}{} & \multicolumn{1}{c|}{672} & {0.117} & \textbf{0.242} & 0.142 & 0.290 & 0.136 & 0.290 & \textbf{0.113} & 0.257 &\multicolumn{1}{|c}{0.277} & 0.431 & \multicolumn{1}{r}{0.200} & \multicolumn{1}{r}{0.367} \\
%ETTm2 24
\midrule
\multicolumn{1}{c|}{\multirow{5}[2]{*}{\begin{sideways}ETTm2\end{sideways}}} & \multicolumn{1}{c|}{24} & \textbf{0.022} & \textbf{0.099} & 0.038 & 0.139 & 0.045 & 0.151 & 0.027 & 0.112 &\multicolumn{1}{|c}{0.036} & 0.141 & \multicolumn{1}{r}{0.048} & \multicolumn{1}{r}{0.153} \\
%ETTm2 48
\multicolumn{1}{c|}{} & \multicolumn{1}{c|}{48} & \textbf{0.045} & \textbf{0.149} & 0.069 & 0.194 & 0.080 & 0.201 & 0.054 & 0.159 & \multicolumn{1}{|c}{0.069} & 0.200 & \multicolumn{1}{r}{0.063} & \multicolumn{1}{r}{0.191} \\
%ETTm2 96
\multicolumn{1}{c|}{} & \multicolumn{1}{c|}{96} & \textbf{0.068} & \textbf{0.189} & 0.089 & 0.225 & 0.094 & 0.229 & 0.072 & 0.196 & \multicolumn{1}{|c}{0.095} & 0.240 & \multicolumn{1}{r}{0.129} & \multicolumn{1}{r}{0.265} \\
%ETTm2 288
\multicolumn{1}{c|}{} & \multicolumn{1}{c|}{288} & {0.160} & {0.272} & 0.161 & 0.306 & 0.155 & 0.309 & \textbf{0.153} & \textbf{0.307} & \multicolumn{1}{|c}{0.211}& 0.367 & \multicolumn{1}{r}{0.208} & \multicolumn{1}{r}{0.352} \\
%ETTm2 672
\multicolumn{1}{c|}{} & \multicolumn{1}{c|}{672} & {0.249} & {0.334} & 0.201 & 0.351 & 0.197 & 0.352 & \textbf{0.183} & \textbf{0.329} & \multicolumn{1}{|c}{0.267} & 0.417 & \multicolumn{1}{r}{0.222} & \multicolumn{1}{r}{0.377} \\
\midrule
%exchange 24
\multicolumn{1}{c|}{\multirow{5}[2]{*}{\begin{sideways}Exchange\end{sideways}}} & \multicolumn{1}{c|}{24} & \textbf{0.027} & \textbf{0.128} & 0.033 & 0.142 & 0.082 & 0.227 & 0.028 & 0.128 &\multicolumn{1}{|c}{0.103} & 0.262 & \multicolumn{1}{r}{-} & \multicolumn{1}{r}{-} \\
%exchange 48
\multicolumn{1}{c|}{} & \multicolumn{1}{c|}{48} & \textbf{0.049} & \textbf{0.169} & 0.059 & 0.191 & 0.116 & 0.268 & 0.048 & 0.169 &\multicolumn{1}{|c}{0.121} & 0.283 & \multicolumn{1}{r}{-} & \multicolumn{1}{r}{-} \\
%exchange 168
\multicolumn{1}{c|}{} & \multicolumn{1}{c|}{168} & \textbf{0.158} & \textbf{0.314} & 0.180 & 0.340 & 0.275 & 0.411 & 0.161 & 0.319 &\multicolumn{1}{|c}{0.168} & 0.337 & \multicolumn{1}{r}{-} & \multicolumn{1}{r}{-} \\
%exchange 336
\multicolumn{1}{c|}{} & \multicolumn{1}{c|}{336} & \textbf{0.382} & \textbf{0.488} & 0.465 & 0.533 & 0.579 & 0.582 & 0.399 & 0.497 &\multicolumn{1}{|c}{1.672} & 1.036 & \multicolumn{1}{r}{-} & \multicolumn{1}{r}{-} \\
%exchange 720
\multicolumn{1}{c|}{} & \multicolumn{1}{c|}{720} & {1.600} & 1.016 & \textbf{1.357} & \textbf{0.931} & 1.570 & 1.024 & 1.639 & 1.044 & \multicolumn{1}{|c}{2.478} & 1.310 & \multicolumn{1}{r}{-} & \multicolumn{1}{r}{-} \\
\midrule
%WTH 24
\multicolumn{1}{c|}{\multirow{5}[2]{*}{\begin{sideways}Weather\end{sideways}}} & \multicolumn{1}{c|}{24} & {0.098} & {0.214} & \textbf{0.096} & 0.215 & 0.102 & 0.221 & \textbf{0.096} & \textbf{0.213} &\multicolumn{1}{|c}{0.117} & 0.251 & \multicolumn{1}{r}{0.109} & \multicolumn{1}{r}{0.217} \\
%WTH 48
\multicolumn{1}{c|}{} & \multicolumn{1}{c|}{48} & \textbf{0.136} & \textbf{0.260} & 0.140 & 0.264 & 0.139 & 0.264 & 0.138 & 0.262 &\multicolumn{1}{|c}{0.178} & 0.318 & \multicolumn{1}{r}{0.143} & \multicolumn{1}{r}{0.269} \\
%WTH 168
\multicolumn{1}{c|}{} & \multicolumn{1}{c|}{168} & \textbf{0.120} & \textbf{0.328} & 0.207& 0.335 & 0.198 & 0.328 & 0.207 & 0.334 &\multicolumn{1}{|c}{0.266} & 0.398 & \multicolumn{1}{r}{0.188} & \multicolumn{1}{r}{0.319} \\
%WTH 336
\multicolumn{1}{c|}{} & \multicolumn{1}{c|}{336} & {0.221} & {0.349} & 0.231 & 0.360 & 0.215 & 0.347 & 0.230 & 0.356 &\multicolumn{1}{|c}{0.197} & 0.416 & \multicolumn{1}{r}{\textbf{0.192}} & \multicolumn{1}{r}{\textbf{0.320}} \\
%WTH 720
\multicolumn{1}{c|}{} & \multicolumn{1}{c|}{720} & 0.235 & {0.365} & 0.233 & 0.365 & 0.219 & 0.353 & 0.242 & {0.370} & \multicolumn{1}{|c}{0.359} & 0.466 & \multicolumn{1}{r}{\textbf{0.198}} & \multicolumn{1}{r}{\textbf{0.329}} \\
\midrule
\multicolumn{2}{c}{Avg.} & \textbf{0.169} & \textbf{0.268} & 0.176 & 0.289 & 0.201 & 0.313 & 0.174 & 0.176 & 0.309 & 0.385 & - & -  \\
\bottomrule

\end{tabular}
    }
  
\caption{Univariate forecasting results of ETT datasets. The best results are highlighted in bold. $L$ denotes the predicted horizons of datasets. The performances are measured in mean-squared error (MSE) and mean-absolute error (MAE).}
\label{tab:datasetuni}%
\end{table}%

%%%%%%%%%%%%%%%%%%%%%%%%%%%%%%%%%%%%%%%%%%%%%%%%%%%%%%%%%%%%%%%%%%%%%%%%%%%%%%%
%%%%%%%%%%%%%%%%%%%%%%%%%%%%%%%%%%%%%%%%%%%%%%%%%%%%%%%%%%%%%%%%%%%%%%%%%%%%%%%

\end{document}